\documentclass[letterpaper, 10 pt, conference]{ieeeconf}  

\IEEEoverridecommandlockouts                              

\usepackage{times}
\usepackage{epsfig}
\usepackage{graphicx}
\usepackage{amsmath}
\usepackage{amssymb}
\usepackage{multirow}
\usepackage{graphics} 
\usepackage{mathptmx} 
\usepackage{subfigure}
\usepackage{url}
\usepackage{epstopdf}

\title{Visual Odometry with an Event Camera Using\\ Continuous Ray Warping and Volumetric Contrast Maximization
}

\author{Yifu Wang, Jiaqi Yang, Xin Peng, Peng Wu, Ling Gao, Kun Huang, Jiaben Chen, Laurent Kneip
\thanks{Y. Wang, J. Yang, X. Peng, P. Wu, L. Gao, K. Huang, J. Chen and L. Kneip are with the Mobile Perception Lab of School of Information Science and Technology at the ShanghaiTech University, Shanghai, China. E-mail: see http://mpl.sist.shanghaitech.edu.cn}%
}

\begin{document}

\maketitle
\thispagestyle{empty}
\pagestyle{empty}

\begin{abstract}
We present a new solution to tracking and mapping with an event camera. The motion of the camera contains both rotation and translation, and the displacements happen in an arbitrarily structured environment. As a result, the image matching may no longer be represented by a low-dimensional homographic warping, thus complicating an application of the commonly used Image of Warped Events (IWE). We introduce a new solution to this problem by performing contrast maximization in 3D. The 3D location of the rays cast for each event is smoothly varied as a function of a continuous-time motion parametrization, and the optimal parameters are found by maximizing the contrast in a volumetric ray density field. Our method thus performs joint optimization over motion and structure. The practical validity of our approach is supported by an application to AGV motion estimation and 3D reconstruction with a single vehicle-mounted event camera. The method approaches the performance obtained with regular cameras, and eventually outperforms in challenging visual conditions.
\end{abstract}

\section{Introduction}
Vision-based localization and mapping is an important technology with many applications in robotics, intelligent transportation, and intelligence augmentation. Although several decades of active research have lead to a certain level of maturity, we keep facing challenges in scenarios with high dynamics, low texture distinctiveness, or challenging illumination conditions \cite{fuentes2015visual,cadena2016past}. Event cameras---also called dynamic vision sensors---present an interesting alternative in this regard, as they pair HDR with high temporal resolution. The advantages and challenges of event-based vision are well explained by the original work of Brandli~et~al.~\cite{brandli2014240} as well as the recent survey by Gallego~et~al.~\cite{gallego2019event}.

Previous works have employed time-continuous parametrizations of image warping functions. Based on the assumption that events are pre-dominantly triggered by high-gradients edges in the image, the optimal image warping parameters will cause the events to warp onto a sharp edge-map in a reference view called the Image of Warped Events (IWE). The optimal warping parameters are hence found by maximizing contrast in the IWE. Various reward functions to evaluate contrast have been presented and analysed in the recent works of Gallego~et~al.~\cite{gallego2018unifying,gallego2019focus} and Stoffregen~and~Kleeman~\cite{stoffregen2019event1}, and successfully used for solving a variety of problems with event cameras such as optical flow~\cite{zhu2017event,gallego2018unifying,stoffregen2018simultaneous,DBLP:journals/corr/abs-1809-08625,zhu2019unsupervised,zhu2018ev}, segmentation~\cite{stoffregen2018simultaneous,stoffregen2019event,mitrokhinevent}, 3D reconstruction~\cite{rebecq2018emvs,zhu2018realtime,zhu2019unsupervised,DBLP:journals/corr/abs-1809-08625}, and motion estimation~\cite{gallego2017accurate,gallego2018unifying,peng2020globally,peng2021globally,liu2020globally}. The main problem with the construction of the IWE is that it relies on a low-dimensional image-to-image warping function, which---in the case of both translational and rotational displacements---is only possible if the model is homographic or if knowledge about the depth of the scene is prior available. 

\begin{figure}[t]
  \centering
  \includegraphics[width=0.70\linewidth]{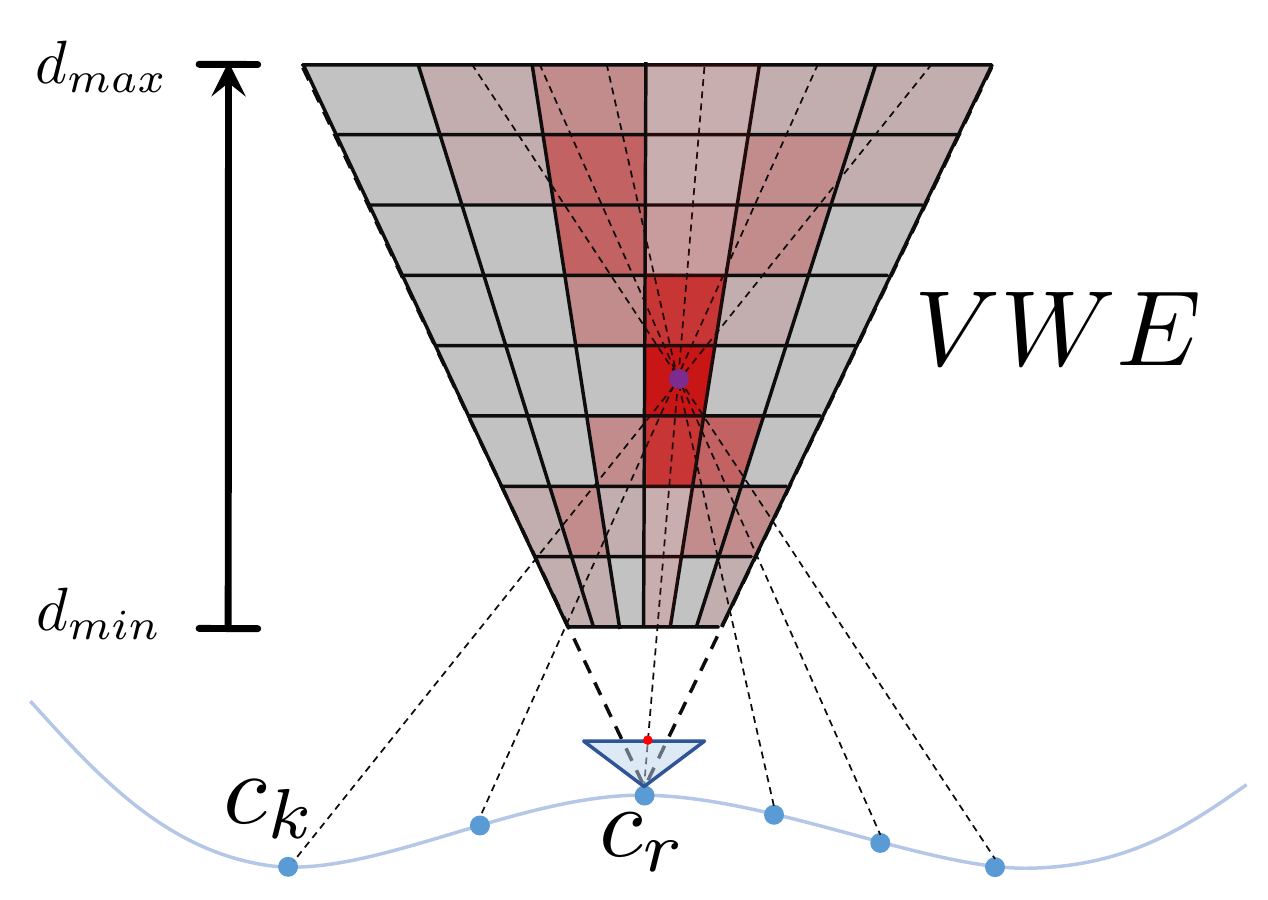}
  \vspace{-0.2cm}
  \caption{Volume of Warped Events: Events are transformed into rays that are warped in space based on a continuous time trajectory model. We evaluate the ray density in a volume in front of a reference view, and maximize its contrast as a function of the continuous motion parameters.}
  \vspace{-0.6cm}
  \label{fig:volume}
\end{figure}

Past solutions to vision-based localization and mapping therefore looked at alternative solution attempts. Note that there are lots of works on the localization and mapping problems individually, a listing of which would go beyond the scope of this introduction. Here we only focus on combined solutions to both problems that use only a single event camera, and that are able to handle combined rotational and translational displacements in unknown, arbitrarily structured environments. There are surprisingly few works that solve this problem, which is proof of its difficulty. The first solution to this problem is given by Kim et al. \cite{kim2016real}, who propose a complex framework of three individual filters. Results are limited to small scale environments and small, dedicated motions. A geometric attempt is given by Rebecq et al. \cite{rebecq2016evo}, who present a combination of a tracker and their ray-density based structure extraction method EMVS \cite{rebecq2018emvs}. However, the framework alternates between the tracking and mapping solutions, which leaves open questions as to how to safely bootstrap the system. Zhu et al. \cite{zhu2019neuromorphic} finally present a promising learning-based approach. It does however depend on vast amounts of training data, and provides no guarantees of optimality or generality.

Our work makes the following contributions:
\begin{itemize}
\item We perform contrast maximization in 3D. Using a time-continuous trajectory model, the 3D location of the landmarks corresponding to events is modelled by time-continuous ray warping in space, and the optimal motion parameters are found by maximizing contrast within a volumetric ray density field.
\item Our method is the first to perform joint optimization over motion and structure for event cameras exerting both translational and rotational displacements in an arbitrarily structured environment.
\item We successfully apply our framework to Autonomous Ground Vehicle (AGV) motion estimation with a forward facing event camera. We prove that using only an event camera, we can provide good quality, continuous visual localization and mapping results able to compete with regular camera alternatives, especially as visual conditions degrade.
\end{itemize}

\section{Contrast Maximization}
\label{sec:CMfoundations}

We are given a set of $N$ events $\mathcal{E} = \{e_{k}\}_{k=1}^{N}$ happening over a certain time interval, where each event $e_k = \{ \mathbf{x}_k, t_k, b_k \}$ is defined by its image location $\mathbf{x}_k=[\begin{matrix} x_k & y_k \end{matrix}]^T$, timestamp $t_k$, and polarity $b_k$.  Note that the set is ordered, meaning that if $\mathcal{E}=\{ \ldots, e_{i}, \ldots, e_{j}, \ldots \}$, then $t_i \leq t_j$. We furthermore assume that image warping during the entire time interval can be parametrized as a continuous-time function of a certain parameter vector $\boldsymbol\theta$, and define the warping function $\mathbf{x'}_k=\mathbf{W}(\mathbf{x}_k,t_{k}|\boldsymbol{\theta})$ that warps an event with location $\mathbf{x}_k$ and timestamp $t_k$ into a reference view at $t_{r}$. 


Gallego et al. recently propose a unifying framework for solving motion estimation problems with event cameras \cite{gallego2018unifying}. If the motion is estimated correctly, events which are triggered by the same point will be accumulated by the same pixel in the reference view, and the resulting Image of Warped Events (IWE) will therefore become a sharp edge map. The question is how the accumulation is done, and how the sharpness of the IWE is characterised. 

Gallego et al. propose to optimize the alignment of the events by maximizing the contrast in the IWE. Formally, the IWE at point $\mathbf{x}$ is defined by $I(\mathbf{x}|\mathbf{\theta}) = \sum_{k = 1}^{N} e^{-\frac{\|\mathbf{x} - \mathbf{x'}_k\|^2}{2\sigma}}$,
and it is evaluated discretely for each pixel centre location. While the application of a Gaussian kernel makes sure that events which are closer to a certain pixel contribute more than events that are further away, it also makes sure that the IWE and its contrast remain smooth functions of the motion parameters and thus optimizable through gradient-based methods. According to \cite{stoffregen2019event1} and \cite{gallego2019focus}, the contrast or sharpness of the IWE may finally be evaluated using one of several possible focus loss functions. Here we use the perhaps most common one, given by the IWE variance $f_{Var}(I) = \frac{1}{N_{p}}\sum_{i,j}(I(\mathbf{x}_{ij}|\boldsymbol\theta) - \mu_{I})^2$.
$\mu_{I}$ is the mean value of $I$, $N_{p}$ the number of pixels in $I$, and $i$ and $j$ are indices that loop through all the rows and columns of the IWE. As shown in the heat maps of \cite{gallego2019focus}, the highest variance of the IWE gives the highest contrast location, and thus the optimal motion parameters causing the best alignment of the warped events. 

The framework allows us to tackle several important motion estimation problems for event-based vision, such as optical flow estimation, motion segmentation, or pure rotational motion estimation. However, note that for an arbitrary point to be warped into the reference view, the warping must either be homographic, or the parameter vector $\theta$ must contain the depth for each event at the time it was captured. Both are rather restrictive towards general camera motion estimation in arbitrary environments. Current state-of-art contrast maximization methods are therefore only able to handle a particular set of problems such as motion in front of a plane, or pure rotation.

\section{Volumetric Contrast Maximization using Ray Warping}

Let us now proceed to our main contribution, which consists of extending the idea of contrast maximization into 3D, a technique that will enable us to handle situations in which we perceive non-planar environments under arbitrary motion and with no priors on the depth of events. Our main idea is illustrated in Figure \ref{fig:volume}. We introduce a continuous-time camera trajectory model as done in Furgale et al. \cite{furgale2015continuous}, which parametrizes both the position and the orientation of the sensor as a smooth, continuous function of time. For a given event, we may then use its timestamp to extrapolate the position and orientation of the event camera at the time the event was captured. Combined with the normalised spatial direction of the event inside the camera frame, each event can be translated into a spatial ray for which the starting point and orientation depend on the continuous trajectory parameters. Rather than evaluating the density of points for pixels in the image, we then propose to evaluate the density of rays at discrete locations in a volume in front of a reference view. We denote this volumetric density field the \textit{Volume of Warped Events (VWE)}. The intuition is analogous to the IWE: the assumption is that there is a limited number of spatial (appearance or geometric) edges that will cause sufficiently large gradients in the image. Under the optimal motion parameters, the rays of the events will therefore intersect along those spatial edges and cause maximum \textit{ray density} in those regions. In other words, the optimal motion parameters may be found by maximizing the contrast in the VWE. The important question is again given by how to express the ray density in the VWE.

The structure of the VWE field is inspired by the space-sweeping approach of \cite{rebecq2018emvs} et al., who propose to estimate 3D structure regardless of explicit data associations and photometric information by finding local maxima in a spatial ray density field. However, their method assumes known camera poses, and they use an alternative camera tracking scheme in their previous work \cite{rebecq2016evo}. To the best of our knowledge, we are the first to propose the maximization of the contrast in the volumetric ray density field, and thus implicitly perform joint optimization over the continuous camera trajectory parameters and the 3D structure.
\subsection{Continuous Ray Warping}
\label{sec:creating VWE}
Suppose our event camera is pre-calibrated and camera-to-image as well as image-to-camera transformation functions $\pi(\cdot)$ and $\pi^{-1}(\cdot)$ are given. The latter transforms image locations into spatial directions in the camera frame by $\mathbf{f}_k = \pi^{-1}(\mathbf{x}_k)$.
In terms of the extrinsics, the trajectory of the camera is kept general for now and simply represented by a minimal, time-continuous, smoothly varying 6-vector $\mathbf{s}(t|\boldsymbol{\theta}) = \left[ \begin{matrix} \mathbf{t}(t|\boldsymbol{\theta}) \\ \mathbf{q}(t|\boldsymbol{\theta}) \end{matrix} \right]$,
where $\boldsymbol{\theta}$ still represents a set of continuous motion parameters, $\mathbf{t}$ the position of the camera expressed in a world frame, and $\mathbf{q}$ its orientation as a Rodriguez vector. Note that the dimensionality of $\boldsymbol{\theta}$ is left unspecified for now. However, as will be shown in Section \ref{sec:ackermann}, it may indeed have only one or two parameters for certain special types of planar displacements. Besides their inherently smooth property, the continuous-time trajectory model has the obvious ability of being able to register information coming from temporally dense sampling sensors, such as event cameras. The transformation from camera to world at time $t$ is given by $\mathbf{T}(t|\boldsymbol{\theta}) =\left[\begin{matrix}\mathbf{R}(\mathbf{q}(t|\boldsymbol{\theta})) & \mathbf{t}(t|\boldsymbol{\theta}) \\ \mathbf{0}^\intercal & 1 \end{matrix}\right]$.
With reference to Figure \ref{fig:volume}, $c_k$ represents the camera frame at time $t_k$ where a certain event $e_k$ has been captured. The absolute pose of the frame at the time of capturing $e_k$ is given by $\mathbf{T}_{wk} = \mathbf{T}(t_k|\boldsymbol{\theta})$. Now let $c_r$ be the reference frame in which we define the projective sampling volume for the VWE. The absolute pose of $c_r$ is given by $\mathbf{T}_{wr} = \mathbf{T}(t_r|\boldsymbol{\theta})$. The relative transformation is finally given as
\begin{equation}
    \mathbf{T}_{rk} = \left[\begin{matrix}\mathbf{R}_{r}(t_k|\boldsymbol{\theta}) & \mathbf{t}_{r}(t_k|\boldsymbol{\theta}) \\ \mathbf{0}^\intercal & 1 \end{matrix}\right] = \mathbf{T}_{wr}^{-1}\mathbf{T}_{wk}.
\end{equation}

Finally, let $\lambda$ represents the unknown depth along the ray. Any point on the ray seen from the reference view can be parametrized by $\mathbf{p}_k(\lambda) = \lambda\mathbf{R}_{r}(t_k|\boldsymbol{\theta})\mathbf{f}_k+ \mathbf{t}_{r}(t_k|\boldsymbol{\theta})$.
\subsection{VWE and Spatial Contrast Maximization}
\label{sec:ray warping}

\begin{figure}[t!]
  \centering
  \includegraphics[width=0.60\linewidth]{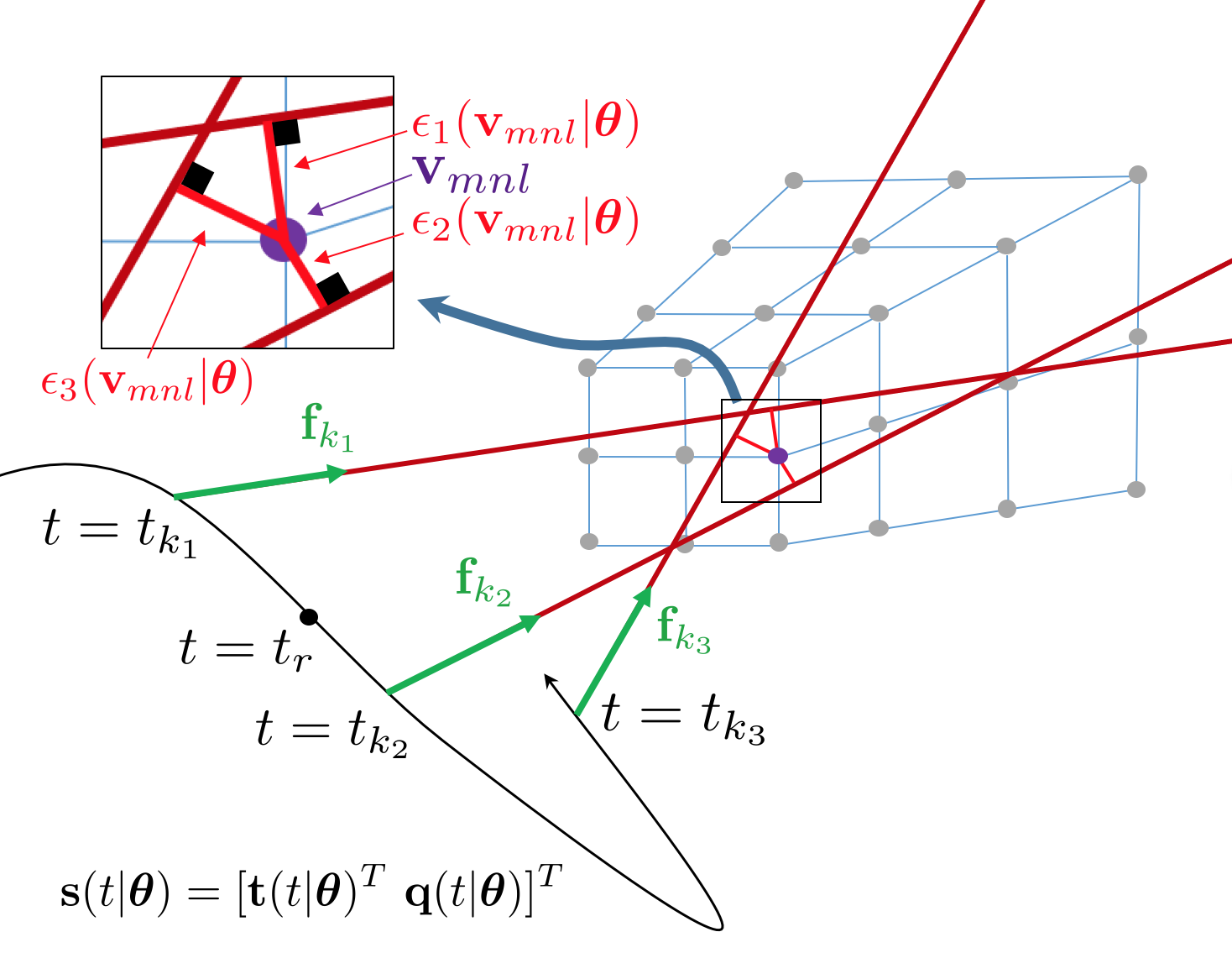}
  \vspace{-0.4cm}
  \caption{Warped rays with object space distances for an example voxel $\mathbf{v}_{mnl}$.}
  \label{fig:object_dist}
  \vspace{-0.7cm}
\end{figure}

We are now going back to our question of how to express the ray density in the VWE. The VWE is defined in a volumetric, projective sampling grid as illustrated in Figure \ref{fig:object_dist}. Let $\mathbf{v}$ be the centre of a voxel. The density of the rays in a voxel is now expressed as a function of the orthogonal distance between the voxel centre $\mathbf{v}$ (expressed in the reference view) and each individual ray. This spatial point-to-line distance is also called the \textit{object space distance}, and it is given by
\begin{equation}
    \epsilon_k^r(\mathbf{v}|\boldsymbol{\theta}) = \|(\mathbf{I} - \mathbf{V}_k)(\mathbf{R}^\intercal_r(t_k|\boldsymbol\theta)(\mathbf{v} - \mathbf{t}_r(t_k|\boldsymbol{\theta})))\|,
\end{equation}
where we have used the rotation $\mathbf{R}^\intercal_r(t_k|\boldsymbol\theta)$ and translation $-\mathbf{R}^\intercal_r(t_k|\boldsymbol\theta)\mathbf{t}_r(t_k|\boldsymbol\theta)$ to transform the voxel centre $\mathbf{v}$ into the camera viewpoint at time $t_k$, and $(\mathbf{I} - \mathbf{V}_k)=(\mathbf{I} - \frac{\mathbf{f}_k\mathbf{f}_k^T}{\mathbf{f}_k^T\mathbf{f}_k})$ is the householder matrix to project this point onto the normal plane of the observation direction $\mathbf{f}_k$. An example of object space distances for one voxel is indicated in Figure \ref{fig:object_dist}.

\begin{figure}[b!]
  \centering
  \vspace{-0.6cm}
  \includegraphics[width=0.7\linewidth]{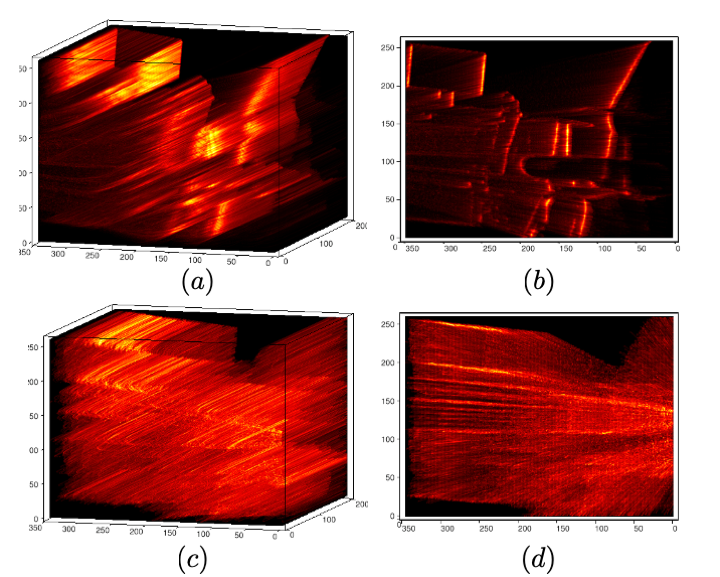}
  \vspace{-0.4cm}
  \caption{Volumetric ray density fields for correct ((a) and (b)) and wrong ((c) and (d)) motion parameters.}
  \label{fig:VWE}
\end{figure}
Supposing that we have $N$ events, the final VWE is again given in smooth form by applying a Gaussian kernel and summing up the object space distances of every event with respect to the voxel centre $\mathbf{v}$
\begin{equation}
	V^r(\mathbf{v}|\boldsymbol\theta) = \sum_{k = 1}^N e^{-\frac{\epsilon_k^r(\mathbf{v}|\boldsymbol\theta)^2}{2\sigma}}.
	\label{eq:kernel}
\end{equation}
The standard deviation $\sigma$ of the Gaussian kernels is actually not constant, but chosen as a function of the depth of the corresponding voxel from the centre of the reference view.

The final optimization objective is given by maximizing the variance of the VWE, which expresses the sharpness of the edges reflected in the volumetric density field
\begin{equation}
	    f_{Var}(V^r) = \frac{1}{N_{v}}\sum_{m,n,l}(V^r(\mathbf{v}_{mnl}|\boldsymbol\theta) - \mu_{V^r})^2.
\end{equation}
$\mu_{V^r}$ is the mean value of $V^r$, $N_{v}$ the number of voxels in the entire volume, and $m$, $n$, and $l$ now iterate through the voxels in the volume. Figure \ref{fig:VWE} visualizes an example VWE for wrong and correct motion parameters. For correct motion parameters (cf. \ref{fig:VWE}(a) and \ref{fig:VWE}(b)), the density field presents higher values and more contrast than for wrong motion parameters (cf. \ref{fig:VWE}(c) and \ref{fig:VWE}(d)).

\subsection{Global optimization over longer trajectories}

\begin{figure}[b!]
    \centering
    \vspace{-0.4cm}
    \includegraphics[width=0.9\columnwidth]{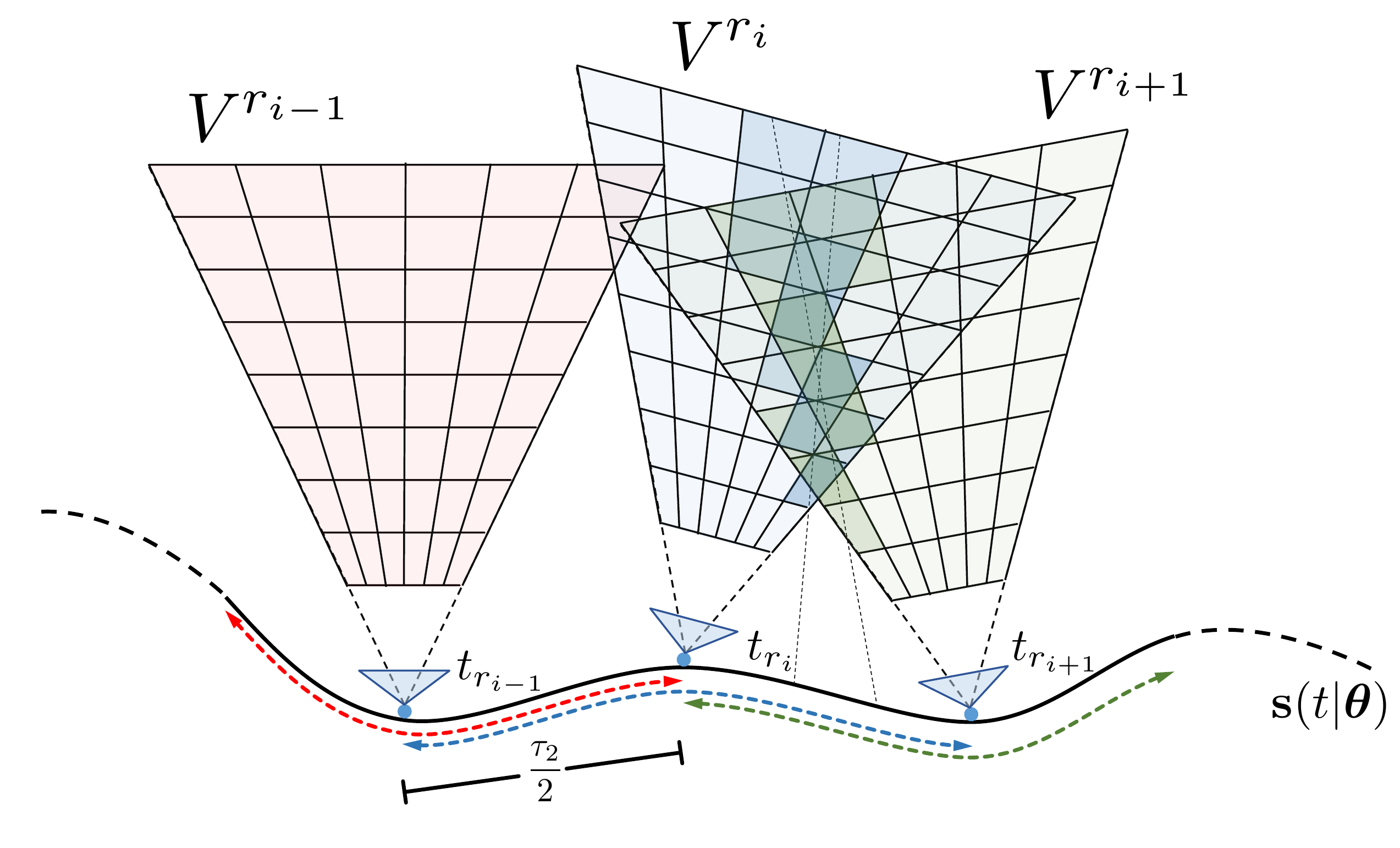}
    \vspace{-0.4cm}
    \caption{Global optimization over multiple reference volumes. The volumes may have spatial overlap. There is an individual time span $[t_{r_i}-\frac{1}{2}\tau_2:t_{r_i}+\frac{1}{2}\tau_2]$ associated with each reference volume $V^{r_i}$ from which events will be considered (marked by the red, blue and green arrows). The time-spans may have temporal overlap. Two events may hence both appear in two distinct density fields, which reinforces scale propagation in the optimization.}
    \label{fig:longertraj}
\end{figure}

We perform global optimization by simultaneously maximizing the contrast in multiple VWEs cast from neighbouring reference views. Let $\{t_{r_1},\ldots,t_{r_M}\}$ be the time instants at which individual VWEs are placed. For simplicity, the time instants are regularly spaced such that $t_{r_{i+1}}-t_{r_i}=\tau_1$. We furthermore define time intervals $[t_{r_i}-\frac{\tau_2}{2}:t_{r_i}+\frac{\tau_2}{2}]$ for each corresponding field $V^{r_i}$, which define the subset of events that will be used for registration in that reference view. More specifically, event $e_k$ is used in $V^{r_i}$ if $t_k\in[t_{r_i}-\frac{\tau_2}{2}:t_{r_i}+\frac{\tau_2}{2}]$. The overall global optimization objective becomes
\vspace{-0.2cm}
\begin{equation}
    \underset{\boldsymbol\theta}{\max}
    \sum_{i=1}^{M}\frac{1}{N_{v}}\sum_{m,n,l}(V^{r_i}(\mathbf{v}_{mnl}|\boldsymbol\theta) - \mu_{V^{r_i}})^2,\nonumber
    \label{eq:finalEq}
\end{equation}
\vspace{-0.4cm}
\begin{equation}
	\text{where } V^{r_i}(\mathbf{v}|\boldsymbol\theta) = \sum_{e_k \in \mathcal{E}_{t_{r_i}-\frac{1}{2}\tau_2}^{t_{r_i}+\frac{1}{2}\tau_2}} e^{-\frac{\epsilon_k^{r_i}(\mathbf{v}|\boldsymbol\theta)^2}{2\sigma}},
\end{equation}
and $\mathcal{E}_{t_{r_i}-\frac{1}{2}\tau_2}^{t_{r_i}+\frac{1}{2}\tau_2}$ is defined as the subset of all the events $e_k$ for which $t_k\in[t_{r_i}-\frac{\tau_2}{2}:t_{r_i}+\frac{\tau_2}{2}]$. The global optimization strategy is depicted in Figure \ref{fig:longertraj}. Note that $\tau_1$ may be chosen such that neighbouring volumes are overlapping, and $\tau_2$ may be chosen such that events are considered in more than just a single volumetric density field (i.e. $\tau_2 > \tau_1$). These choices guarantee that the implicit graph behind this optimization problem is well connected and effects such as scale propagation take place.

\section{Application to AGV with a Forward-Facing Event Camera}
\label{sec:ackermann}

We evaluate our method on a planar Autonomous Ground Vehicle~(AGV) on which we mount a single forward-facing event camera. Many solutions for regular, monocular cameras exist, such as simple relative pose solvers \cite{nister04} or full visual SLAM frameworks \cite{mur2017orb}. The application of an event camera promises strong advantages in situations of high motion dynamics or---as shown in this work---challenging illumination conditions. Our motion estimation framework is divided into two sub-parts, a front-end module that initializes motion over shorter segments, and a back-end module that refines the estimate over larger-scale sequences. Both will be introduced after a short overview of the framework.



\subsection{Framework overview}

The complete VO system is designed based on the above VWE method. There are two main modules in the pipeline. The front-end initialization module groups the events into sufficiently small subsets such that the motion on these subsets can be locally approximated using a simplified first-order constant velocity model. Furthermore, the front-end performs contrast maximization using a single VWE only. After a sufficient number of events and initial relative displacements have been accumulated, our method then proceeds to the back-end optimization part. The latter initializes a larger-scale, smooth, continuous-time trajectory model and executes the multi-volume optimization outlined in \eqref{eq:finalEq}.

\subsection{Front-end single-frame optimization}
%
%

For the local approximation of the motion, we use a parametrization that is inspired by \cite{scaramuzza2009real} and \cite{Huang2019vehicle}. Based on the assumptions of a driftless, non-holonomic platform, and locally constant velocities, the continuous motion of the planar vehicle may be approximated to lie on an arc of a circle to which the heading of the vehicle remains tangential. This motion model is also known as the Ackermann motion model, and the centre of the circle is commonly referred to as the Instantaneous Centre of Rotation (ICR). The model has only two degrees of freedom, which largely simplifies the geometry of the problem. It is given by the forward velocity $v$ and the rotational velocity $\omega$.

Using the convention and equations from \cite{Huang2019vehicle}, the relative transformation from a frame at time $t_k$ to a nearby reference frame at time $t_r$ is given by
\begin{equation}
\begin{aligned}
	& \mathbf{R}_r^v(t_k|\boldsymbol\theta) = \left[\begin{matrix}
							\cos\omega(t_k-t_r) & -\sin\omega(t_k-t_r)  & 0 \\
							\sin\omega(t_k-t_r)   & \cos\omega(t_k-t_r)  & 0  \\
	 						0 & 0 & 1 
				\end{matrix}\right]\\
	& \mathbf{t}_r^v(t_k|\boldsymbol\theta)= \frac{v}{\omega}\left[\begin{matrix}
							1-\cos(\omega(t_k-t_r))\\
							\sin(\omega(t_k-t_r))\\
	 						0
				\end{matrix}\right].
\end{aligned}
\label{equ:acker_constraint} 
\end{equation}
Given that scale is unobservable, we fix the forward velocity $v$ to the configured speed of the vehicle (correct scale propagation is taken into account in the later global optimization scheme). As a result, the local motion initialization scheme over a single volume has only 1-DoF, and the parameter vector becomes $\boldsymbol\theta=\omega$. Note furthermore that the original Ackermann model requires the camera to be mounted in the centre of the non-steering axis, which---in practice---hardly ever is the case. We therefore add the extrinsic calibration parameters $\mathbf{R}_{vc}$ and $\mathbf{t}_{vc}$ which transform points back-and-forth between the camera and the vehicle reference frames. 
The reader is invited to read up \cite{Huang2019vehicle} to see more foundations of the Ackermann motion model. Note that the variance of the VWE is a function of our unique degree of freedom $\omega$, and the motion parameters can thus be efficiently solved by local gradient-based optimization methods once a rough initial guess is given.



%
%
\subsection{Back-end multi-frame optimization}
\label{sec:bsplineBA}


The front-end obviously estimates the motion over short time periods, only, and furthermore relies on the approximation of locally constant velocities and a circular arc trajectory. We add a global back-end optimization over the entire trajectory which relies on a more general model for representing smooth planar motion. We use a two-dimensional, $p$-th degree B-spline curve
\begin{equation}
  \mathbf{c}_{2\times 1}(t) = \sum_{i=0}^{n} N_{i,p}(t) \mathbf{p}_i ,\qquad a\leq t\leq b,
\end{equation}
where the $\{\mathbf{p}_i\}$ stand for the $n+1$ two-dimensional control points that control the shape of the smooth, planar trajectory, and the $\{N_{i,p}(t)\}$ are the known $p$th-degree B-spline basis functions. The reader is invited to read up \cite{piegl2012nurbs} and \cite{furgale2015continuous} to see the foundations of B-splines and example applications. Here we only focus on establishing the link to our smooth camera pose functions used in the optimization objective \eqref{eq:finalEq}.

The parameter vector $\boldsymbol\theta$ may be defined as the stacked control points of the spline expressed by $\boldsymbol\theta = [\mathbf{p}_0^T \text{ } \ldots \text{ } \mathbf{p}_n^T]^T$.
The spline directly models the position in the plane, so we easily obtain $\mathbf{t}^v(t|\boldsymbol\theta) = \left[\begin{matrix}\mathbf{c}(t) \\ 0\end{matrix}\right]$.
For planar motion, the orientation is given by a pure rotation about the vertical axis, and we furthermore exploit the fact that for driftless non-holonomic vehicles, the heading of the vehicle remains tangential to the trajectory. If the heading of the vehicle is still defined as the $y$ axis, and the $z$ axis points vertically upwards, the orientation of the vehicle is finally given as
\begin{equation}
    \mathbf{R}^v(t|\boldsymbol\theta)=
    \left[\begin{matrix}
      \left[\begin{matrix}0 & 1 \\ -1 & 0 \end{matrix}\right] \mathbf{\dot{c}}(t) & \mathbf{\dot{c}}(t) & \mathbf{0} \\ 0 & 0 & 1
    \end{matrix}\right].
\end{equation}
Note that only the temporal basis functions depend on time, and that $\mathbf{\dot{c}}(t)$ therefore also is a spline-based function of the same control points. 
The control point vector is initialized from the approximate trajectory given by the front end using the spline curve approximation given by the automatic knots spacing algorithms (9.68) and (9.69) of \cite{piegl2012nurbs}.

\section{Implementation and validation}

In this section, we briefly introduce implementation details of our method and then test our algorithm on multiple both synthetic and real datasets. We assess both accuracy and quality of the estimated trajectories, as well as the quality of the implicitly modelled 3D structure.

\subsection{Implementation details}

We utilize the event back-projection approach proposed in \cite{rebecq2018emvs} to efficiently find the neighbouring voxels of a spatial ray. The details of this algorithm can be found in Section 7.1 of \cite{rebecq2018emvs}. We furthermore use a simple gradient-ascent scheme to solve our volumetric contrast maximization problems. Especially in (\ref{equ:acker_constraint}), the fixation of the forward velocity $v$ leaves the angular velocity $\omega$ as the only unknown parameter, thus making the front-end constraint a uni-variate problem. Finally, in order to recover the implicitly modelled 3D structure of the environment, we simply reuse the Event-based Multi-View Stereo (EMVS) method from \cite{rebecq2018emvs}.

\subsection{Experiment setup}

In order to demonstrate the performance of our algorithm, we apply it to both synthetic and real datasets. In the synthetic case, we use large-scale outdoor sequences from the KITTI benchmark \cite{Geiger2013IJRR} and convert the image sequences into event data by using the method of Gehrig et al. \cite{Gehrig2020CVPR}. The datasets are fully calibrated and contain images captured by a forward-looking camera mounted on a vehicle driving through a city. Experiments on real data are conducted by collecting several small-scale indoor sequences with a DAVIS346 event camera. The camera is mounted forward-facing on a turtlebot. It has a resolution of 346$\times$260, and captures RGB images in parallel to the events.

We compare our approach against traditional camera alternatives. Our current implementation focuses on non-holonomic planar motion, which is why we use the 1-point RANSAC algorithm for Ackermann motion \cite{scaramuzza2009real} as a solid baseline algorithm for the regular camera alternative. We also let our method compete against an established alternative from the open-source community: ORB-SLAM \cite{mur2017orb}. Note that we rescale all monocular, scale-invariant results to align as well as possible with groundtruth, which we obtain from the original KITTI datasets or an Opti-track system.

It should be noted that a direct comparison against alternative event-based VO/SLAM projects is difficult for several reasons. Til date, there are no open-source implementations and we are the first to even evaluate a monocular, event-based pipeline on a popular, established benchmark sequence. Furthermore, as stated in Section III. D of \cite{rebecq2016evo} and Section 3.5 of \cite{kim2016real}, the few existing alternatives either depend strongly on the quality of an initial 3D map (cf. \cite{rebecq2016evo}), or suffer from slowly converging depth estimates (cf. \cite{kim2016real}). As shown in their experiments, they therefore require hovering motion in front of the same scene to provide sufficient time for the mapping back-end to converge. In contrast, our method performs joint optimization of trajectory and structure in near real-time, and thus successfully handles the continuous forward-exploration scenario.

\subsection{Experiment on synthetic data}
\label{sec: synthetic-data}
To prove the effectiveness of our method---which denote \textbf{ETAM}---, we apply it to synthetic sequences generated from the KITTI benchmark datasets \cite{Geiger2013IJRR}. These datasets represent a fairly normal usecase without high motion dynamics or challenging illumination. We use the publicly available tool proposed by \cite{Gehrig2020CVPR} to convert the regular videos into events streams. We compare our method against two alternatives, which is the state-of-the-art \textbf{ORB-SLAM} algorithm \cite{mur2017orb} and the classical 1-point Ransac algorithm---denoted \textbf{1pt}---for planar motion \cite{scaramuzza2009real}. The evaluation is performed on sequences \textit{VO-00} and \textit{VO-07}.

The qualitative performance is illustrated in Figure \ref{fig:realResult}(a). All algorithms successfully process the sequences without any gross errors, and our system is slightly less accurate than \textbf{ORB-SLAM} on these high quality datasets. We furthermore believe that the decrease in performance is mostly explained by the approximate motion model, which ignores the slight pitch angle variations that could result from uneveness of the ground surface. Furthermore, we perform on par with \textbf{1pt}, which also relies on a non-holonomic planar motion model. To the best of our knowledge, this result is the first to demonstrate a monocular event camera solution that returns comparable results to regular camera alternatives.

\subsection{Experiment on real data}
\label{sec: real-data}

\begin{figure}[b]
\centering
\vspace{-0.4cm}
\includegraphics[width=0.40\linewidth]{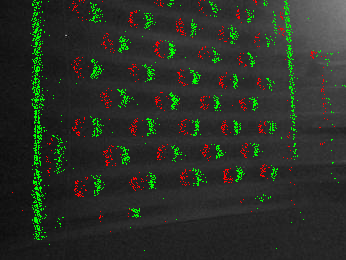}
\includegraphics[width=0.40\linewidth]{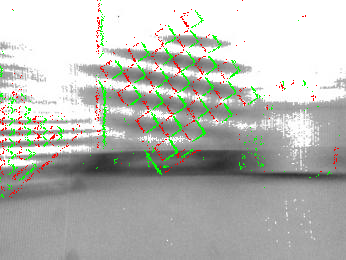} 
\caption{\textit{Challenging illumination conditions}. Regular frames suffer from poor contrast or motion blur when lights are off (left) or on (right), respectively. Events in turn preserve the visual information of the structure.}
\label{fig:blurryimage}
\end{figure}

\begin{figure}[t]
\centering
\subfigure[synthetically generated outdoor sequences]{
	\includegraphics[width=0.445\linewidth]{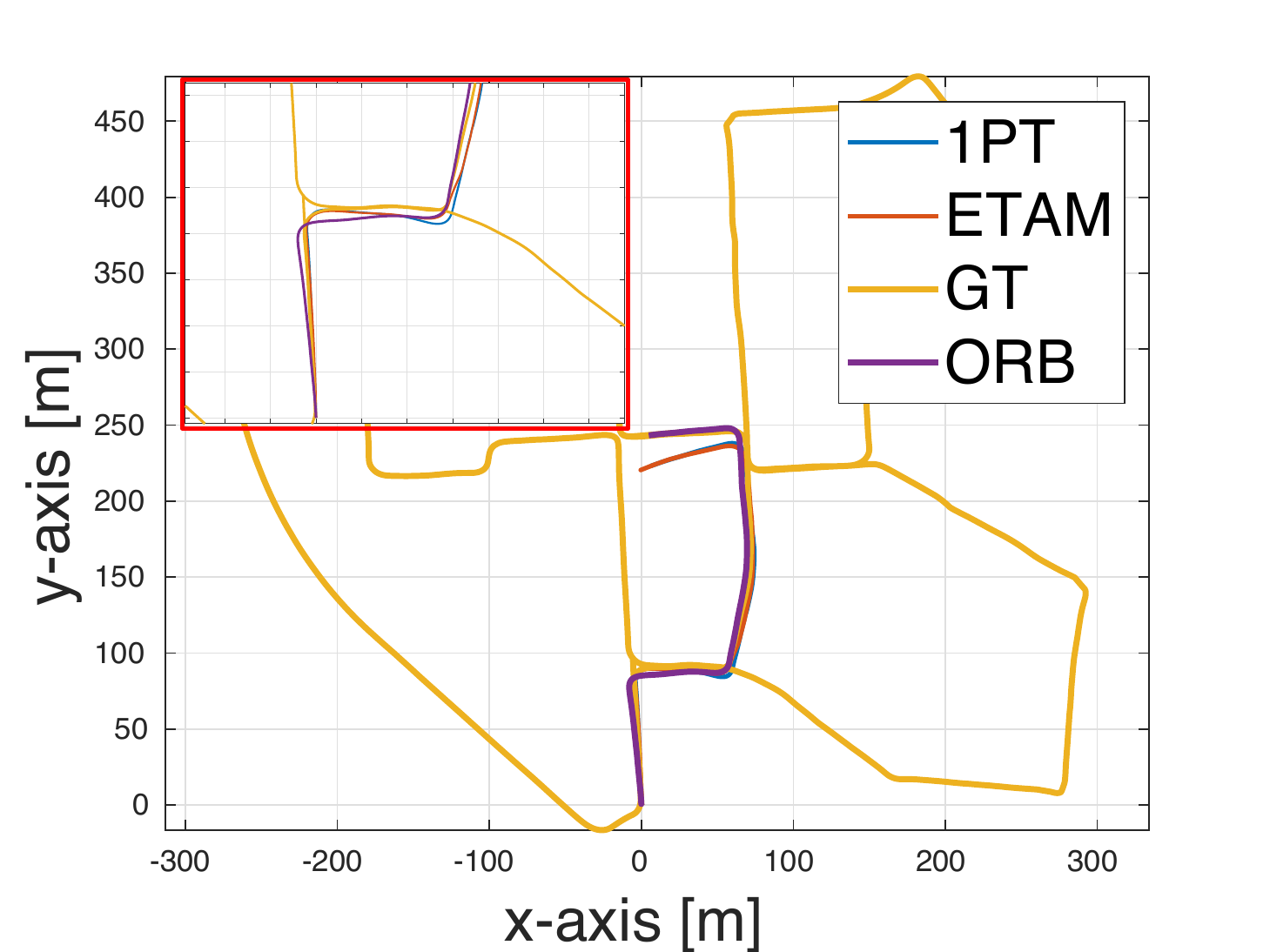}
	\includegraphics[width=0.445\linewidth]{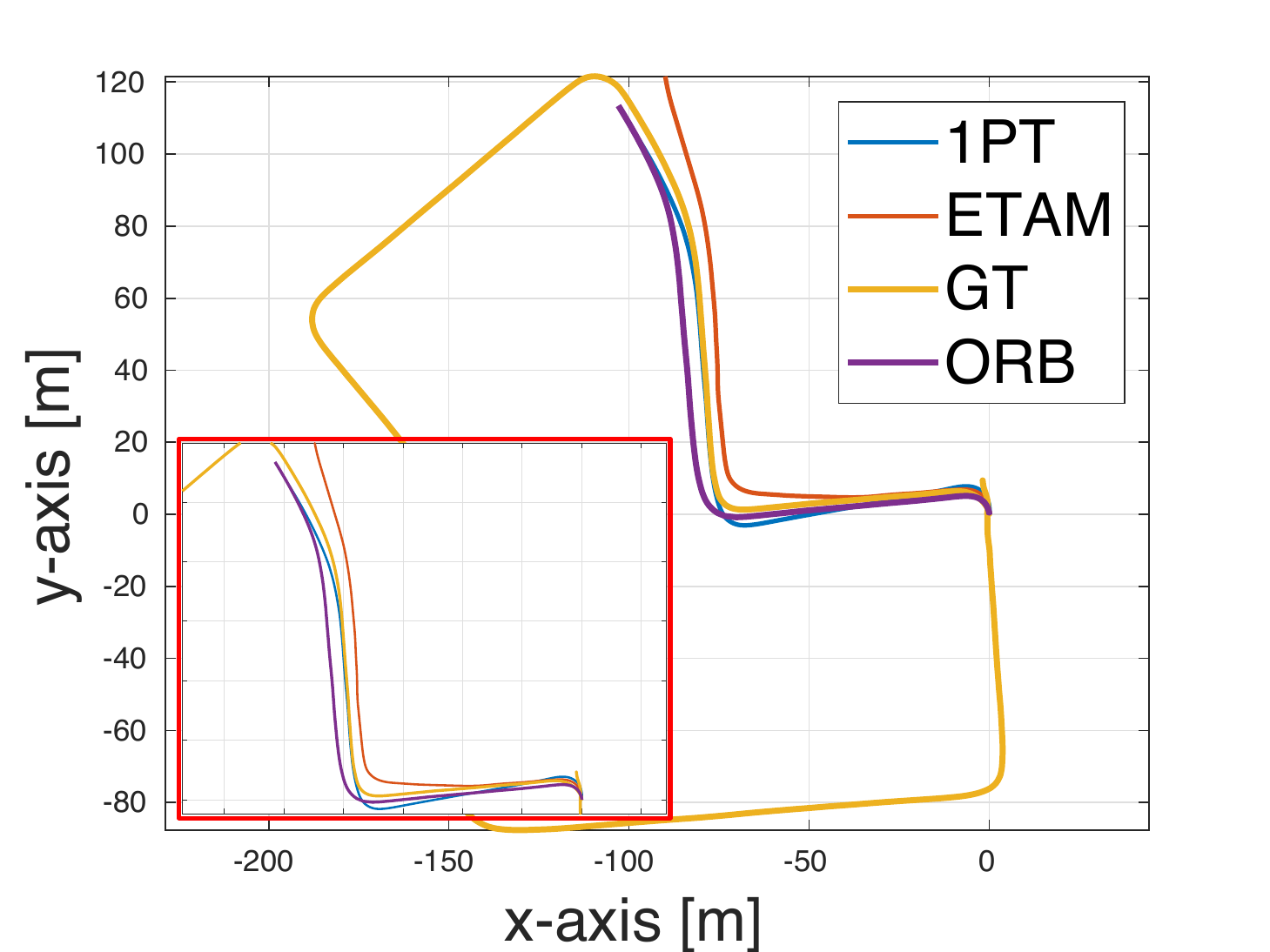} }
\subfigure[real data indoor sequences]{
	\includegraphics[width=0.425\linewidth]{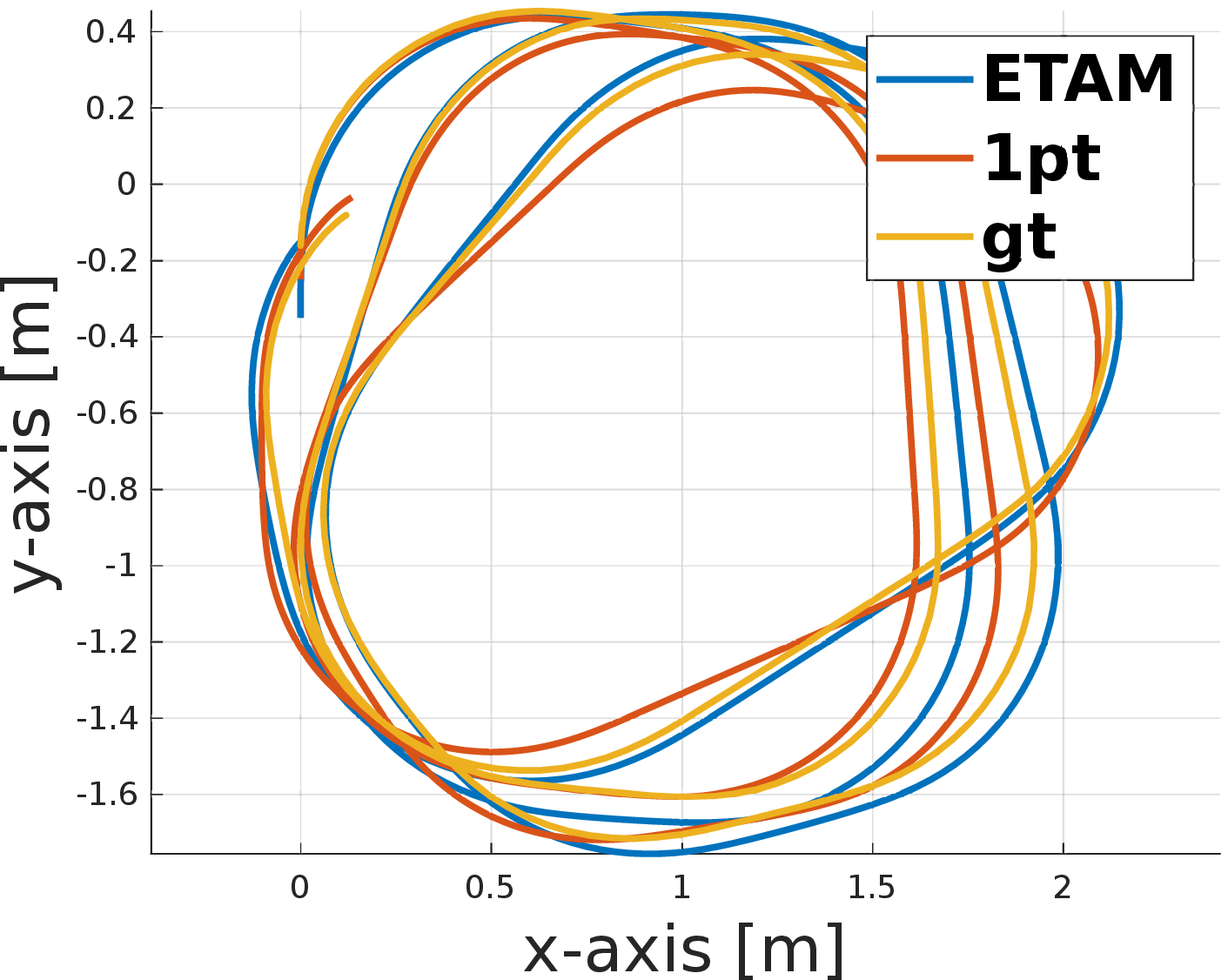}
	\includegraphics[width=0.425\linewidth]{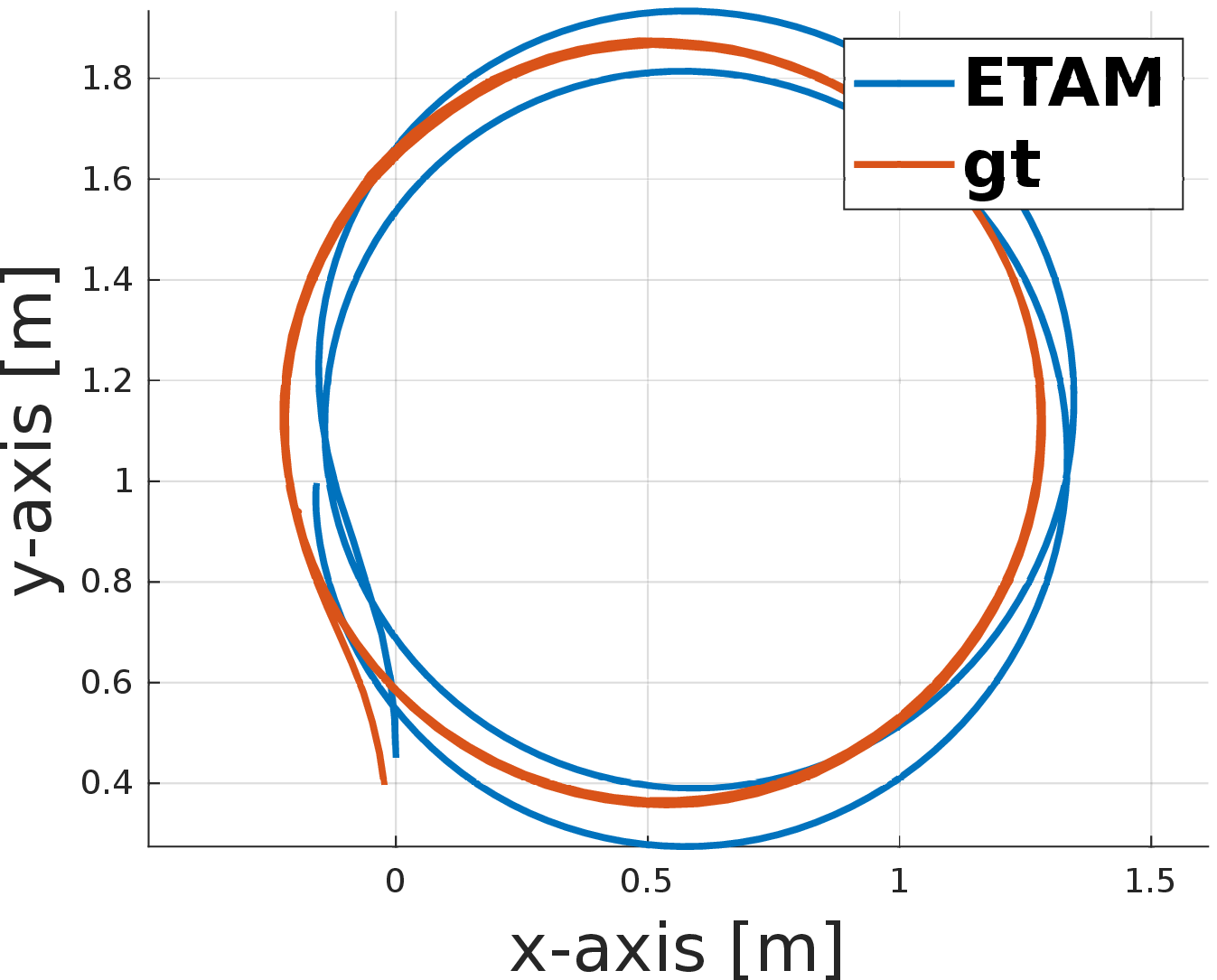} }
\caption{Results for both our method and 1pt-RANSAC on long outdoor trajectories (top) and indoor sequences (bottom). The indoor sequences are captured under normal (left) or challenging illumination conditions (right).}
\label{fig:realResult}
\vspace{-0.6cm}
\end{figure}

In order to demonstrate the performance of our algorithm on real data, we collect further sequences with a DAVIS346 event camera. The datasets are captured indoors to simulate different illumination conditions and capture groundtruth via Opti-track. We first apply them to two shorter sequences in which the camera follows an either circular (\textit{Circle}) or purely translational trajectory (\textit{Str}). Next, we perform a test on a much longer sequence with more complex motion (\textit{Long2}). While the first three sequences are recorded under good illumination conditions, we conclude with another sequence with varying lighting conditions by toggling external illumination while the dataset is recorded (\textit{HDR}).

\textbf{ORB-SLAM} proves to be fragile when applied to our indoor sequences. The images have low resolution and the proximity of the structure as well as fast vehicle rotations furthermore induce large frame-to-frame disparities, ultimately causing ORB-SLAM to break in such forward-exploration scenarios. We therefore assess the performance by a quantitative comparison of relative pose errors between \textbf{ETAM} and \textbf{1pt}. Results for all sequences are summarized in Table \ref{table:realtest}. It shows the root-mean-square or median of all deviations between estimated and groundtruth short-term relative rotation and translation displacements. Note that---in order to minimize the impact of unobservable scale---the error of the relative translation is evaluated by considering only the direction. Furthermore, errors are assessed per time, as it is clear that larger intervals may lead to more drift. We therefore employ the unit $deg/s$ for both rotational and translational errors. The best performance is always highlighted in bold.

\renewcommand\arraystretch{1.25}
\begin{table}[ht!]
\centering
\vspace{0.2cm}
\caption{Accuracy on different sequences. Unit: $[deg/s]$.}
\vspace{-0.2cm}
\begin{tabular}{|l|c|c|c|c|}
\hline
\multirow{2}{*}{method}         & \multicolumn{4}{c|}{Circular motion}                                                            \\ \cline{2-5} 
                                & Rmse($\mathbf{R}$)   & Median($\mathbf{R}$)   & Rmse($\mathbf{t}$)   & Median($\mathbf{t}$)     \\ \hline
\multirow{1}{*}{\textbf{1pt}}   & 2.4526               & 2.3330                 & 0.5427               & \textbf{0.0296}          \\ \hline
\textbf{ETAM}                   & \textbf{1.3275}      & \textbf{0.6443}        & \textbf{0.2826}      & 0.0322                   \\  \hline
\end{tabular}

\vspace{0.05cm}
\begin{tabular}{|l|c|c|c|c|}
\hline
\multirow{2}{*}{method}         & \multicolumn{4}{c|}{Purely translational motion}                                                \\ \cline{2-5} 
                                & Rmse($\mathbf{R}$)   & Median($\mathbf{R}$)   & Rmse($\mathbf{t}$)   & Median($\mathbf{t}$)     \\ \hline
\multirow{1}{*}{\textbf{1pt}}   & \textbf{0.6997}      & \textbf{0.5300}        & 0.5179               & 0.0369                   \\ \hline
\textbf{ETAM}                   & 0.9769               & 0.6334                 & \textbf{0.4637}      & \textbf{0.0235}          \\  \hline
\end{tabular}

\vspace{0.05cm}
\begin{tabular}{|l|c|c|c|c|}
\hline
\multirow{2}{*}{method}         & \multicolumn{4}{c|}{Long trajectory}                                                             \\ \cline{2-5} 
                                & Rmse($\mathbf{R}$)   & Median($\mathbf{R}$)   & Rmse($\mathbf{t}$)   & Median($\mathbf{t}$)     \\ \hline
\multirow{1}{*}{\textbf{1pt}}   & 1.8516               & 1.5829                 & 0.1675               & 0.1718                   \\ \hline
\textbf{ETAM}                   & \textbf{1.6901}      & \textbf{1.3417}        & \textbf{0.1631}      & \textbf{0.1703}          \\  \hline
\end{tabular}

\vspace{0.05cm}
\begin{tabular}{|l|c|c|c|c|}
\hline
\multirow{2}{*}{method}         & \multicolumn{4}{c|}{Challenging illumination conditions}                                        \\ \cline{2-5} 
                                & Rmse($\mathbf{R}$)   & Median($\mathbf{R}$)   & Rmse($\mathbf{t}$)   & Median($\mathbf{t}$)     \\ \hline
\multirow{1}{*}{\textbf{1pt}}   & -                    & -                      & -                    & -                        \\ \hline
\textbf{ETAM}                   & \textbf{1.6042}      & \textbf{0.9093}        & \textbf{0.0686}      & \textbf{0.0084}          \\  \hline
\end{tabular}
\vspace{-0.6cm}
\label{table:realtest}
\end{table}

It can be easily observed that \textbf{ETAM} outperforms \textbf{1pt} on most datasets, and it is able to continuously track entire sequences with high accuracy even as illumination conditions become more challenging. In contrast, regular camera based visual odometry with 1-point RANSAC fails due to poor contrast or motion blur in dark or varying illumination settings (cf. Figure \ref{fig:blurryimage}). Due to the forward-facing arrangement, the purely translational displacement on the other hand triggers much fewer events than trajectories with rotational displacements, hence the slightly inferior performance for this type of motion. Figure \ref{fig:realResult}(b) visualizes top views of complete trajectories for both algorithms and groundtruth (denoted \textbf{gt}). The left figure is from the sequence \textit{Long2}, and the right one is from the sequence \textit{HDR}. Our event-based method is able to work robustly in all challenging conditions. We kindly refer the reader to our supplemental video file for further qualitative results of our method.

\subsection{Computational efficiency}
All experiments are conducted on an Intel Core i7 2.4 GHz CPU. The total cumulative processing time for each sequence is summarized in Table \ref{table:opttime}. It remains below the actual length of each dataset, thus indicating real-time capability.

\begin{table}[h]
\vspace{-0.2cm}
\centering
\caption{Processing time in seconds for the proposed method.}
\vspace{-0.2cm}
\begin{tabular}{|c|c|c|c|c|c|c|}
\hline
 & \textit{Circle} & \textit{Str}& \textit{Long1} & \textit{Long2} & \textit{HDR}\\
\hline
Dataset length & 14.0s & 10.4s & 43.3s & 40.4s & 17.9s\\
\hline
Processing time & 8.6s & 8.5s & 35.6s & 24.9s & 13.8s\\
\hline
\end{tabular}
\label{table:opttime}
\vspace{-0.5cm}
\end{table}

\section{CONCLUSIONS}


Our main novelty consists of a single, joint objective that optimizes smooth motion directly from events, without the need of a prior derivation of 3D structure. This is achieved by constructing a volumetric ray density field, in which we then maximize contrast as a function of smooth motion parameters. As a result, the approach is able to bootstrap spatial motion in arbitrarily structured environments. The formulation is tested on the important application of ground vehicle motion estimation, and potential advantages under high dynamic motion or challenging illumination conditions are verified. While this is a highly promising result, our next step consists of extending the operation to more dynamic, full 3D motion, which we believe is possible if using the additional input of an IMU.

{\small
\bibliographystyle{IEEEtran}
\bibliography{root}

\begin{thebibliography}{10}
\providecommand{\url}[1]{#1}
\csname url@samestyle\endcsname
\providecommand{\newblock}{\relax}
\providecommand{\bibinfo}[2]{#2}
\providecommand{\BIBentrySTDinterwordspacing}{\spaceskip=0pt\relax}
\providecommand{\BIBentryALTinterwordstretchfactor}{4}
\providecommand{\BIBentryALTinterwordspacing}{\spaceskip=\fontdimen2\font plus
\BIBentryALTinterwordstretchfactor\fontdimen3\font minus
  \fontdimen4\font\relax}
\providecommand{\BIBforeignlanguage}[2]{{%
\expandafter\ifx\csname l@#1\endcsname\relax
\typeout{** WARNING: IEEEtran.bst: No hyphenation pattern has been}%
\typeout{** loaded for the language `#1'. Using the pattern for}%
\typeout{** the default language instead.}%
\else
\language=\csname l@#1\endcsname
\fi
#2}}
\providecommand{\BIBdecl}{\relax}
\BIBdecl

\bibitem{fuentes2015visual}
J.~Fuentes-Pacheco, J.~Ruiz-Ascencio, and J.~M. Rend{\'o}n-Mancha, ``Visual
  simultaneous localization and mapping: a survey,'' \emph{Artificial
  intelligence review}, vol.~43, no.~1, pp. 55--81, 2015.

\bibitem{cadena2016past}
C.~Cadena, L.~Carlone, H.~Carrillo, Y.~Latif, D.~Scaramuzza, J.~Neira, I.~Reid,
  and J.~J. Leonard, ``Past, present, and future of simultaneous localization
  and mapping: Toward the robust-perception age,'' \emph{IEEE Transactions on
  robotics}, vol.~32, no.~6, pp. 1309--1332, 2016.

\bibitem{brandli2014240}
C.~Brandli, R.~Berner, M.~Yang, S.-C. Liu, and T.~Delbruck, ``A 240$\times$ 180
  130 db 3 $\mu$s latency global shutter spatiotemporal vision sensor,''
  \emph{IEEE Journal of Solid-State Circuits}, vol.~49, no.~10, pp. 2333--2341,
  2014.

\bibitem{gallego2019event}
G.~Gallego, T.~Delbruck, G.~Orchard, C.~Bartolozzi, B.~Taba, A.~Censi,
  S.~Leutenegger, A.~Davison, J.~Conradt, and K.~Daniilidis, ``Event-based
  vision: A survey,'' \emph{IEEE Transactions on Pattern Analysis and Machine
  Intelligence}, 2020.

\bibitem{gallego2018unifying}
G.~Gallego, H.~Rebecq, and D.~Scaramuzza, ``A unifying contrast maximization
  framework for event cameras, with applications to motion, depth, and optical
  flow estimation,'' in \emph{Proceedings of the IEEE Conference on Computer
  Vision and Pattern Recognition}, 2018, pp. 3867--3876.

\bibitem{gallego2019focus}
G.~Gallego, M.~Gehrig, and D.~Scaramuzza, ``Focus is all you need: loss
  functions for event-based vision,'' in \emph{Proceedings of the IEEE
  Conference on Computer Vision and Pattern Recognition}, 2019, pp.
  12\,280--12\,289.

\bibitem{stoffregen2019event1}
T.~Stoffregen and L.~Kleeman, ``Event cameras, contrast maximization and reward
  functions: an analysis,'' in \emph{Proceedings of the IEEE Conference on
  Computer Vision and Pattern Recognition}, 2019, pp. 12\,300--12\,308.

\bibitem{zhu2017event}
A.~Z. Zhu, N.~Atanasov, and K.~Daniilidis, ``Event-based feature tracking with
  probabilistic data association,'' in \emph{2017 IEEE International Conference
  on Robotics and Automation (ICRA)}.\hskip 1em plus 0.5em minus 0.4em\relax
  IEEE, 2017, pp. 4465--4470.

\bibitem{stoffregen2018simultaneous}
T.~Stoffregen and L.~Kleeman, ``Simultaneous optical flow and segmentation
  (sofas) using dynamic vision sensor,'' \emph{2017 Australasian Conference on
  Robotics and Automation (ACRA)}, pp. 52--61, 2017.

\bibitem{DBLP:journals/corr/abs-1809-08625}
\BIBentryALTinterwordspacing
C.~Ye, A.~Mitrokhin, C.~Parameshwara, C.~Ferm{\"{u}}ller, J.~A. Yorke, and
  Y.~Aloimonos, ``Unsupervised learning of dense optical flow and depth from
  sparse event data,'' \emph{CoRR}, vol. abs/1809.08625, 2018. [Online].
  Available: \url{http://arxiv.org/abs/1809.08625}
\BIBentrySTDinterwordspacing

\bibitem{zhu2019unsupervised}
A.~Z. Zhu, L.~Yuan, K.~Chaney, and K.~Daniilidis, ``Unsupervised event-based
  learning of optical flow, depth, and egomotion,'' in \emph{Proceedings of the
  IEEE Conference on Computer Vision and Pattern Recognition}, 2019, pp.
  989--997.

\bibitem{zhu2018ev}
------, ``Ev-flownet: Self-supervised optical flow estimation for event-based
  cameras,'' \emph{arXiv preprint arXiv:1802.06898}, 2018.

\bibitem{stoffregen2019event}
T.~Stoffregen, G.~Gallego, T.~Drummond, L.~Kleeman, and D.~Scaramuzza,
  ``Event-based motion segmentation by motion compensation,'' in
  \emph{Proceedings of the IEEE International Conference on Computer Vision},
  2019, pp. 7244--7253.

\bibitem{mitrokhinevent}
A.~Mitrokhin, C.~Ferm{\"u}ller, C.~Parameshwara, and Y.~Aloimonos,
  ``Event-based moving object detection and tracking. in 2018 ieee,'' in
  \emph{RSJ International Conference on Intelligent Robots and Systems (IROS)},
  2018, pp. 1--9.

\bibitem{rebecq2018emvs}
H.~Rebecq, G.~Gallego, E.~Mueggler, and D.~Scaramuzza, ``Emvs: Event-based
  multi-view stereo—3d reconstruction with an event camera in real-time,''
  \emph{International Journal of Computer Vision}, vol. 126, no.~12, pp.
  1394--1414, 2018.

\bibitem{zhu2018realtime}
A.~Z. Zhu, Y.~Chen, and K.~Daniilidis, ``Realtime time synchronized event-based
  stereo,'' in \emph{European Conference on Computer Vision}.\hskip 1em plus
  0.5em minus 0.4em\relax Springer, 2018, pp. 438--452.

\bibitem{gallego2017accurate}
G.~Gallego and D.~Scaramuzza, ``Accurate angular velocity estimation with an
  event camera,'' \emph{IEEE Robotics and Automation Letters}, vol.~2, no.~2,
  pp. 632--639, 2017.

\bibitem{peng2020globally}
X.~Peng, Y.~Wang, and L.~Kneip, ``Globally optimal event camera motion
  estimation,'' in \emph{Proceedings of the European Conference on Computer
  Vision (ECCV)}, 2020.

\bibitem{peng2021globally}
X.~{Peng}, L.~{Gao}, Y.~{Wang}, and L.~{Kneip}, ``Globally-optimal contrast
  maximisation for event cameras,'' \emph{IEEE Transactions on Pattern Analysis
  and Machine Intelligence}, pp. 1--1, 2021.

\bibitem{liu2020globally}
D.~Liu, A.~Parra, and T.-J. Chin, ``Globally optimal contrast maximisation for
  event-based motion estimation,'' in \emph{Proceedings of the IEEE/CVF
  Conference on Computer Vision and Pattern Recognition}, 2020, pp. 6349--6358.

\bibitem{kim2016real}
H.~Kim, S.~Leutenegger, and A.~J. Davison, ``Real-time 3d reconstruction and
  6-dof tracking with an event camera,'' in \emph{European Conference on
  Computer Vision}.\hskip 1em plus 0.5em minus 0.4em\relax Springer, 2016, pp.
  349--364.

\bibitem{rebecq2016evo}
H.~Rebecq, T.~Horstsch{\"a}fer, G.~Gallego, and D.~Scaramuzza, ``Evo: A
  geometric approach to event-based 6-dof parallel tracking and mapping in real
  time,'' \emph{IEEE Robotics and Automation Letters}, vol.~2, no.~2, pp.
  593--600, 2016.

\bibitem{zhu2019neuromorphic}
D.~Zhu, Z.~Xu, J.~Dong, C.~Ye, Y.~Hu, H.~Su, Z.~Liu, and G.~Chen,
  ``Neuromorphic visual odometry system for intelligent vehicle application
  with bio-inspired vision sensor,'' in \emph{2019 IEEE International
  Conference on Robotics and Biomimetics (ROBIO)}.\hskip 1em plus 0.5em minus
  0.4em\relax IEEE, 2019, pp. 2225--2232.

\bibitem{furgale2015continuous}
P.~Furgale, C.~H. Tong, T.~D. Barfoot, and G.~Sibley, ``Continuous-time batch
  trajectory estimation using temporal basis functions,'' \emph{The
  International Journal of Robotics Research}, vol.~34, no.~14, pp. 1688--1710,
  2015.

\bibitem{nister04}
D.~Nist\'er, O.~Naroditsky, and J.~Bergen, ``Visual odometry,'' in \emph{{IEEE}
  Conf. Comput. Vis. Pattern Recog. (CVPR)}, Washington, DC, USA, 2004, pp.
  652--659.

\bibitem{mur2017orb}
R.~Mur-Artal and J.~D. Tard{\'o}s, ``{ORB-SLAM2}: An open-source slam system
  for monocular, stereo, and rgb-d cameras,'' \emph{IEEE Transactions on
  Robotics}, vol.~33, no.~5, pp. 1255--1262, 2017.

\bibitem{scaramuzza2009real}
D.~Scaramuzza, F.~Fraundorfer, and R.~Siegwart, ``Real-time monocular visual
  odometry for on-road vehicles with 1-point ransac,'' \emph{2009 IEEE
  International Conference on Robotics and Automation}, pp. 4293--4299, 2009.

\bibitem{Huang2019vehicle}
K.~Huang, Y.~Wang, and L.~Kneip, ``Motion estimation of non-holonomic ground
  vehicles from a single feature correspondence measured over n views,'' in
  \emph{Proceedings of the IEEE Conference on Computer Vision and Pattern
  Recognition}.\hskip 1em plus 0.5em minus 0.4em\relax IEEE, 2019, pp.
  12\,706--12\,715.

\bibitem{piegl2012nurbs}
L.~Piegl and W.~Tiller, \emph{The NURBS book}.\hskip 1em plus 0.5em minus
  0.4em\relax Springer Science \& Business Media, 2012.

\bibitem{Geiger2013IJRR}
A.~Geiger, P.~Lenz, C.~Stiller, and R.~Urtasun, ``Vision meets robotics: The
  kitti dataset,'' \emph{International Journal of Robotics Research (IJRR)},
  2013.

\bibitem{Gehrig2020CVPR}
D.~Gehrig, M.~Gehrig, J.~Hidalgo-Carri\'o, and D.~Scaramuzza, ``Video to
  events: Recycling video datasets for event cameras,'' in \emph{{IEEE} Conf.
  Comput. Vis. Pattern Recog. (CVPR)}, June 2020.

\end{thebibliography}
}
\end{document}